\documentclass[runningheads]{llncs}

\usepackage{eccv}

\usepackage{eccvabbrv}

\usepackage{graphicx}
\usepackage{booktabs}

\usepackage[accsupp]{axessibility}  % Improves PDF readability for those with disabilities.
\usepackage[utf8]{inputenc}

\usepackage{hyperref}

\usepackage{orcidlink}

\begin{document}

\title{RTE-FM-Dehazer: Radiative Transfer Equation Inspired Flow Matching for Real-World Image Dehazing} 

\titlerunning{RTE-FM-Dehazer}

\author{
Chenfeng Wei\inst{1,5}\orcidlink{0000-0003-0560-3049} \and
Chun Wang\inst{2}\orcidlink{0009-0007-2290-910X} \and
Boyang Zhao\inst{3}\orcidlink{0009-0009-2970-6617} \and
Si Zuo\inst{4}\orcidlink{0009-0002-1538-2083} \and
Shenhong Wang\inst{1}$^{,*}$\orcidlink{0009-0005-7085-2465} \and
Chenguang Yang\inst{5,6}$^{,*}$\orcidlink{0000-0001-5255-5559}
}

\authorrunning{C.~Wei et al.}

\institute{%
  \textsuperscript{1}Xi'an Jiaotong-Liverpool University, China\quad
  \textsuperscript{2}Mashang Consumer Finance Co. Ltd, China \quad
  \textsuperscript{3}Tsinghua University, China \quad
  \textsuperscript{4}Hunan University, China \quad
  \textsuperscript{5}University of Liverpool, UK \quad
  \textsuperscript{6}The Hong Kong Polytechnic University, Hong Kong SAR, China \quad
  \textsuperscript{*}\,Corresponding authors: 
  \email{Shenhong.Wang@xjtlu.edu.cn}, 
  \email{Chenguang.Yang@liverpool.ac.uk}
}
\maketitle

% 正文从这里开始
\begin{abstract}

Single-image dehazing aims to recover a clear scene from a hazy image and is generally formulated as an image-to-image translation task; however, it faces two limitations. Its performance depends heavily on the haze-formation priors embedded in the model. Prevailing methods adopt the Atmospheric Scattering Model (ASM), whose assumptions of single scattering and homogeneous media are often violated, leading to residual haze and color drift. Moreover, large-scale real hazy/clear pairs are impractical to collect, and existing synthesis approaches fail to reproduce the full complexity of natural haze. To address these issues, we present RTE-FM-Dehazer, a novel dehazing approach, together with a scalable data pipeline. Unlike the ASM, the Radiative Transfer Equation (RTE) jointly accounts for both scattering and absorption, naturally accommodating the non-homogeneous, multiple-scattering media that characterize real hazy scenes. Motivated by the structural similarity between the RTE diffusion-absorption term and the ODE in flow matching, we introduce a diffusion-absorption regularizer derived from a reduced RTE, to steer the flow matching trajectory at each step. Next, leveraging modern vision–language models, we build an automated pipeline and release P-HAZE, a dataset of 50\,000 realistic hazy/clear pairs. Extensive evaluations demonstrate that RTE-FM-Dehazer, trained solely on P-HAZE, effectively eliminates artifacts like residual haze and color drift, exhibits strong cross-domain generalization, and achieves leading results on five real-world dehazing benchmarks. Code and data are available at \url{https://github.com/vincentweikey/RTE-FM-Dehazer}.

\end{abstract}

\section{Introduction}
\label{sec:intro}

\begin{figure}[!htbp]
  \centering
  \includegraphics[width=\linewidth]{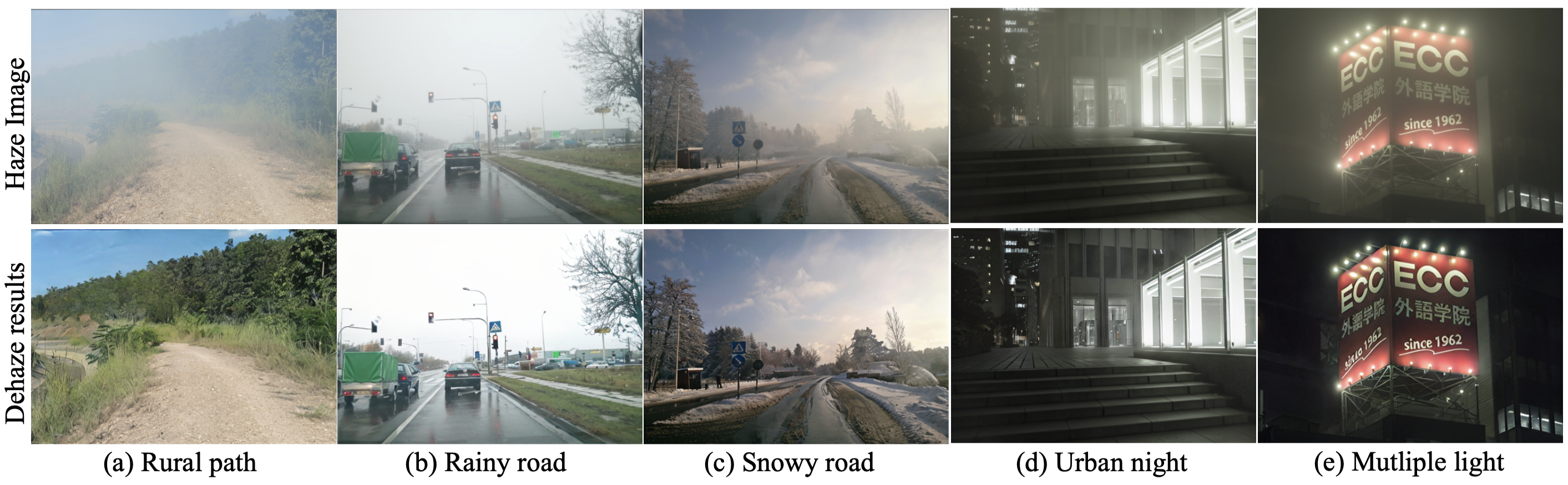}
  \caption{Our \textbf{RTE-FM-Dehazer} efficiently removes spatially-varying haze under complex weather and scenes while preserving original colors and fine textures.}
  \label{fig:overview}
\end{figure}

% background & research field

Haze severely degrades visual quality, posing challenges for real-world applications such as autonomous driving, remote sensing, and outdoor scene analysis. In the real world, haze is spatially uneven and light is multiply scattered, leaving robust dehazing still an active research area. Dehazing is an ill-posed problem; current methods typically impose the atmospheric scattering model(ASM)\cite{ASM} as a physical prior to regularize the solution.

\begin{equation}
I(\mathbf{x})=T(\mathbf{x})\,J(\mathbf{x})+\bigl(1-T(\mathbf{x})\bigr)A,
\end{equation}
where $T(\mathbf{x})\!\in\![0,1]$ is the pixel-wise transmission, $J(\mathbf{x})$ is the clean radiance we wish to recover, and $A\!\in\!\mathbb{R}^3$ is the global air-light introduced by suspended particles.
Estimating both $T$ and $J$ from a single RGB image is inherently under-constrained. Early methods using hand-crafted priors: Dark-Channel statistics~\cite{dark_channel}, Color-Line distributions~\cite{color_lines}, Haze-Line clusters~\cite{haze_line}, or Color-Attenuation ratios\cite{color_attenuation} to estimate the transmission map. These cues are elegant, interpretable, and sufficient when the haze is spatially homogeneous, optically thin, and free of diffuse reflection.

In real-world hazy images, however, these assumptions are routinely violated: haze density varies with scene depth, particles exhibit anisotropic scattering, and multiple scattering dominates. Under such complex conditions, traditional priors break down, leaving residual fog, severe color casts, and over-enhanced artifacts. Recognizing these limitations, recent studies have shifted to learning direct domain-to-domain mappings via powerful deep learning architectures including CNNs \cite{Dehazenet,densely_dehaze,Aodnet,FFAnet,DEAnet} and Transformers \cite{dehazeXL,dehaze3D} etc, achieving impressive performance on synthetic benchmarks. However, these data-driven methods heavily rely on idealized datasets: most training pairs are synthesized under a spatially constant-haze assumption, thus limiting generalization to real-world imagery. To address this, attention has shifted toward better data-synthesis strategies \cite{dehazeXL,domain_dehazing}. Building on this momentum, powerful diffusion-based generative models \cite{dehazediff,LHTD} and VQGAN codebook priors \cite{ridcp,IPC-dehaze} have recently been exploited to improve dehazing performance in real-world settings. Nevertheless, these generative methods introduce a new class of artifacts: the stochastic nature of diffusion sampling may drift toward adjacent color regions, while codebook-based retrieval occasionally grafts patches lying outside the original field of view.

% my work and contribute

Single-image dehazing remains challenging due to spatially-varying haze, scarce training pairs, and generative artifacts. We address these jointly: we replace the restrictive ASM with the RTE\cite{RTE} to model scattering and absorption; we regularize flow matching with physics-inspired deterministic trajectories; and we leverage VLMs with geometric alignment to synthesize scalable training data. Radiative transfer in participating media is governed by the RTE~\cite{RTE}:
\begin{equation}
\frac{\mathrm{d}I}{\mathrm{d}s}=
-\kappa_{\mathrm{a}}I+\kappa_{\mathrm{a}}B(T,\nu)
+\kappa_{\mathrm{s}}\int_{\mathbb S^2}p(\hat{\mathbf{s}},\hat{\mathbf{s}}')I(\mathbf x,\hat{\mathbf{s}}',\nu)\mathrm{d}\Omega',
\label{eq:RTE}
\end{equation}
where $I$ is ray specific intensity, $\kappa_{\mathrm a}$ and $\kappa_{\mathrm s}$ are absorption and scattering coefficients, and the integral captures anisotropic multiple scattering. While the RTE strictly governs radiative transfer in participating media, our latent-space formulation appropriates its diffusion-absorption structure as a soft prior: the learned terms approximate attenuation, soft halos, and varying contrast without the restrictive homogeneity assumptions of the ASM.

Rather than performing a single-step inversion, we regard the hazy observation $\mathbf{I}(\mathbf{x})$ as the initial state of a continuous light-transport trajectory whose equilibrium is the haze-free image $\mathbf{J}(\mathbf{x})$.  
Dehazing thus reduces to computing a physically plausible path $\mathbf{I}\!\to\!\mathbf{J}$ locally regularized by the RTE at each step---a requirement ignored by prior deep networks.
Flow matching~\cite{flow_matching} supplies the ideal vehicle: it evolves the image with an ODE whose velocity field $v_\theta$, once learned, permits single-step Euler integration at test time.  
We therefore introduce \textbf{RTE-FM-Dehazer}, a physics-inspired flow matching framework that regularises $v_\theta$ with an RTE-derived velocity term. To avoid scarce real pairs, we prompt a VLM~\cite{qwen2,gemini} to generate hazy counterparts and align them via dense geometry-aware descriptors and sub-pixel homography warping. Qualitative results on diverse scenes are presented in Figure.~\ref{fig:overview}.
Our main contributions are:
\begin{enumerate}
\item Reformulating single-image dehazing as RTE-regularized flow matching improves physical plausibility by aligning the generative trajectory via a physics-inspired prior, leading to more stable and realistic dehazed results across diverse real-world conditions.
\item Removing stochastic noise priors in the flow matching generation process yields a deterministic latent trajectory regularized by a diffusion-absorption term, maintaining the overall color distribution better than random-sampling baselines.
\item We propose a scalable, VLM-driven pipeline that aligns features and synthesizes realistic haze-clean pairs without requiring scarce real-world data.
\end{enumerate}

\section{Related Work}
\label{sec:formatting}

\subsection{Single Image Dehazing }

Early single-image dehazing approaches are grounded in the atmospheric scattering model and aim to invert the haze formation process by estimating the transmission map. He et al.\cite{dark_channel} proposed the Dark Channel Prior, which exploits the observation that at least one color channel of local patches in haze-free images tends to have very low intensity.  Subsequent priors extended this philosophy: the Haze-line Prior\cite{haze_line} clusters colors along 1-D haze-lines, while the Color Attenuation Prior\cite{color_attenuation} models the relationship between scene depth, brightness and saturation.  These handcrafted priors yield impressive results under moderate haze, yet their linear or low-order statistical assumptions break down for highly textured regions, bright-white objects, or non-uniform illumination, yielding color distortions and halo artifacts.

Driven by large-scale synthetic haze datasets \cite{reside,haze4k,dehazeXL} and the success of CNNs in low-level vision, recent dehazing research has predominantly adopted learning-based approaches. Early CNN-based methods such as DehazeNet~\cite{Dehazenet} and AOD-Net~\cite{Aodnet} directly regress the transmission map or the haze-free image from input observations. Subsequent works have introduced architectural enhancements, including multi-scale feature fusion and attention mechanisms, to enlarge receptive fields and disentangle haze related representations. Notably, FFA-Net~\cite{FFAnet} employs channel and pixel attention to emphasize informative features, while DEA-Net~\cite{DEAnet} integrates detail enhanced convolution with content-guided attention for improved texture restoration. More recent Transformer-based models, such as Dehaze3D~\cite{dehaze3D} and DehazeXL~\cite{dehazeXL}, leverage 3D position embeddings and patch tokenization to capture long-range dependencies and global context in high-resolution images. Despite achieving state-of-the-art PSNR and SSIM \cite{SSIM_PSNR} on synthetic benchmarks, these supervised methods often suffer from significant performance degradation in real-world scenarios due to domain shift. This limitation stems from the oversimplified assumptions of synthetic haze generation, which typically employs homogeneous scattering models and neglects the wavelength-dependent, spatially-varying, mixed scattering-absorption characteristics of real-world haze.

\subsection{Dehazing in real-world}

Confronted with the scarcity of paired real hazy/clear imagery and the limited distributional coverage of synthetic datasets, recent work has sought to mitigate the generalization gap through domain adaptation~\cite{domain_dehazing}, unsupervised priors such as the dark-channel~\cite{unsupervised_dark_channel}, and contrastive regularization~\cite{ucl-dehaze,refinednet}. a parallel line of work treats dehazing as a conditional generation problem. Li et al.~\cite{condition_dehazing} first brought conditional GANs to single-image dehazing, learning a deterministic pixel-to-pixel mapping via adversarial supervision; however, such models demand pixel-perfect aligned pairs and tend to yield over-smoothed predictions once the synthetic training distribution is too narrow. Diffusion probabilistic models~\cite{diffusion} relax this determinism by gradually denoising pure noise toward the clean data manifold. Cheng et al.~\cite{dehazediff} instantiate this idea with DehazeDiff, yet the stochastic reverse process remains sensitive to the variance schedule and the number of denoising steps, making the output quality difficult to control in practice. Vector-quantized auto-encoders (VQ-VAE \cite{VQVAE}, VQ-GAN \cite{VQGAN}) provide a compact discrete latent space that preserves high-frequency details while enabling autoregressive or diffusion-based priors. RIDCP \cite{ridcp} and IPC-Dehaze \cite{IPC-dehaze} incorporate a pre-trained VQ-GAN as a structural prior and finetune the decoder for haze-free reconstruction.
While these approaches produce visually plausible dehazing on real scenes, their generative nature is unconstrained by any physical haze-formation model and thus tends to shift global illumination, bias the original color distribution, and introduce extraneous artifacts. We retain the generative pipeline but embed approximations of the RTE at every flow-matching step, eliminating these artifacts while preserving visual realism.

\section{Methodology}
\label{sec:formatting}
\label{sec:rte-fm}

%-------------------------------------------------------------------------
\subsection{Core Framework}

\begin{figure}[!htbp]
  \centering
  \includegraphics[width=1.0\linewidth]{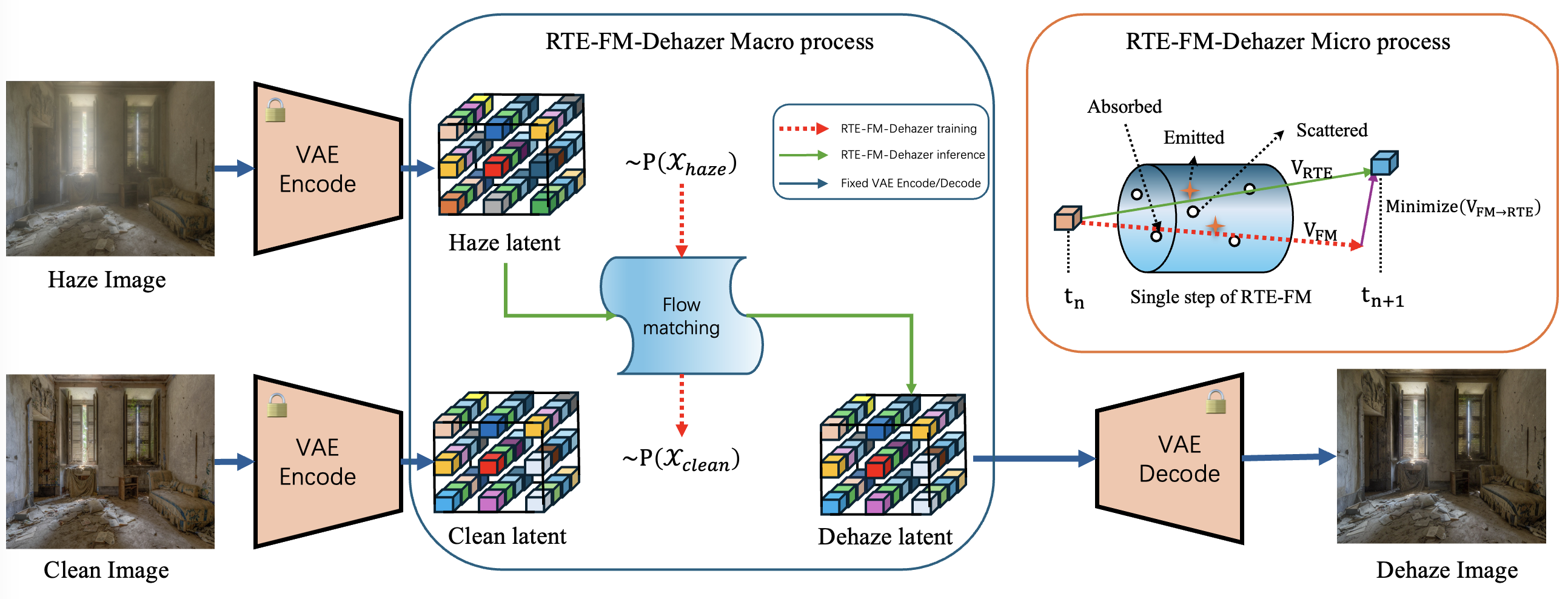}
  \caption{RTE-FM-Dehazer pipeline. In the macro process, a frozen VAE encodes the hazy image into latent space, where flow matching learns a neural velocity field that progressively transports the haze latent toward its clean counterpart. In the micro process, each step refines this velocity so that it aligns with both the data-driven direction and the diffusion–absorption direction estimated by the RTE.
}
  \label{fig:rte-fm-pipline}
\end{figure}

In this section we introduce RTE-FM-Dehazer, a flow matching framework regularized by the Radiative Transfer Equation (RTE). As shown in Figure~\ref{fig:rte-fm-pipline}, flow matching decomposes hazy-to-clean mapping into infinitesimal steps (unlike one-shot VQ-VAE or VQ-GAN), enabling fine-grained injection of RTE-derived absorption and scattering constraints. The network predicts a velocity field that simultaneously satisfies the flow matching objective and aligns with physical dynamics, producing trajectories that approximate valid light transport via ODE integration.

%-------------------------------------------------------------------------
\subsection{RTE-FM-Dehazer Architecture}

\subsubsection{Vanilla Flow Matching Dehazing.}
\label{subsec:fm_euler}

We formulate single-image dehazing as an \emph{Vanilla flow matching} process in the latent space of Stable Diffusion.
Given a hazy image $x_{\text{hazy}}\in\mathbb{R}^{3\times H\times W}$, we first encode it to
$\mathbf{z}_{0}=\mathcal{E}(x_{\text{hazy}})\in\mathbb{R}^{4\times h\times w}$  with the frozen VAE encoder $\mathcal{E}$.
A U-Net is trained to match the \emph{straight-line} velocity
\begin{equation}
\mathcal{L}_{\text{FM}}=\mathbb{E}_{t\sim\mathcal{U}(0,1)}\Bigl[\bigl\|\boldsymbol{v}_{\theta}(\mathbf{z}_{t},t)-(\mathbf{z}_{1}-\mathbf{z}_{0})\bigr\|^{2}\Bigr],
\end{equation}

\begin{equation}
\mathbf{z}_{t}=(1-t)\mathbf{z}_{0}+t\mathbf{z}_{1},
\end{equation}
where $\mathbf{z}_{1}$ is the latent code of the clean target.

At inference we solve the ODE
$\frac{d\mathbf{z}}{dt}=\boldsymbol{v}_{\theta}(\mathbf{z},t)$
by \emph{explicit Euler} with fixed step $\Delta t$:
\begin{equation}
\mathbf{z}_{n+1}=\mathbf{z}_{n}+\Delta t\cdot\boldsymbol{v}_{\theta}(\mathbf{z}_{n},t_{n}).
\end{equation}
Crucially, $\boldsymbol{v}_{\theta}$ outputs a \emph{full-resolution velocity map}, so every spatial location receives its own vector. Consequently \emph{non-uniform haze} is processed with \emph{location-specific speeds}---bright regions, dark regions, and edges are advanced along different directions and magnitudes within the same time step. The method is purely data-driven, learning the velocity field end-to-end without transmission maps or physical priors.

\subsubsection{Flow Matching with RTE Regularizer.}
\label{subsec:fm_rte}
While flow matching regresses to the straight-line velocity $\mathbf{z}_1-\mathbf{z}_0$, it is agnostic to the physical propagation of light in participating media. Accordingly, we start from the diffusion approximation\cite{RTE} of the RTE
\begin{equation}
\frac{\partial u}{\partial t}=D\nabla^{2}u-\mu_{a}u,
\end{equation}
$D\nabla^{2}u$ governs \emph{spatial diffusion}: it redistributes radiative energy from high to low concentration regions, thereby smoothing sharp gradients and suppressing artificial discontinuities in the recovered image. $-\mu_{a}u$ models \emph{medium absorption}: it induces exponential decay of local radiative energy, directly mimicking the attenuating effect of haze or smoke and thus driving the latent state toward a cleaner image manifold.
Where the right-hand side simply expresses the instantaneous rate of change of the density field $u(\mathbf{x},t)$.  In continuum mechanics such a time derivative \emph{is} the Eulerian velocity of the field.  
\begin{equation}
\mathbf{v}_{\text{rte}}(\mathbf{x},t)\triangleq D\nabla^{2}u(\mathbf{x},t)-\mu_{a}u(\mathbf{x},t)
\end{equation}

Drawing an analogy between the physical radiance field $u$ and the latent representation $\mathbf{z}\in\mathbb{R}^{4\times 64\times 64}$, we approximate the RTE dynamics in latent space. 
This approximation is justified because flow matching proceeds via infinitesimal velocity fields, restricting the regularization to the tangent space of the latent manifold where the pretrained decoder acts as a local isometry; consequently, the latent-space diffusion operator induces a transport that is structurally analogous to the physical radiative transfer in the image domain. 
After flipping the sign so that density decrease corresponds to haze removal, this yields:
\begin{equation}
\mathbf{v}_{\text{RTE}}(\mathbf{z})\triangleq -D\nabla^{2}\mathbf{z} + \mu_{a}\mathbf{z}
\end{equation}

\subsubsection{Latent-Space Discretization of RTE Dynamics.}
To evaluate the RTE-derived regularizer efficiently on the latent manifold, we discretize the continuous diffusion operator using the five-point stencil:
\begin{equation}
\resizebox{.9\columnwidth}{!}{%
$\displaystyle
\nabla^{2}\mathbf{z}(x,y)\approx \frac{\mathbf{z}(x+h,y)+\mathbf{z}(x-h,y)+ \mathbf{z}(x,y+h)+\mathbf{z}(x,y-h)-4\mathbf{z}(x,y)}{h^{2}}
$%
}
\end{equation}
where $h$ is the grid spacing in the $64\times 64$ latent space (typically $h=1$), and $(x,y)\in\{1,\dots,H\}\times\{1,\dots,W\}$ denote the integer spatial coordinates indexing the location of latent feature vector $\mathbf{z}(x,y)\in\mathbb{R}^4$ on the feature map. This discrete approximation is evaluated efficiently using convolution with the kernel
\begin{equation}
\mathbf{K}_{\nabla^{2}}=\begin{bmatrix}
0 & 1 & 0\\[2pt]
1 & -4 & 1\\[2pt]
0 & 1 & 0
\end{bmatrix},
\end{equation}
$D$ is estimated by a $1\times 1$ convolution followed by a soft-plus activation. $\mu_{a}=\kappa\langle|\mathbf{z}_{t}|\rangle$ with $\kappa=0.1$ (empirically set to balance the absorption term against the diffusion term in the latent domain), yielding a decoding-free proxy for local absorption strength.

\subsubsection{Regularisation mechanism.}
Flow matching trains $\mathbf{v}_{\theta}$ to reproduce the optimal-transport vector $\mathbf{z}_{1}-\mathbf{z}_{0}$.
We now demand that, at every $t$, $\mathbf{v}_{\theta}$ is simultaneously
(i) aligned with the data-driven direction and
(ii) consistent with the RTE direction.
The combined loss
\begin{equation}
\mathcal{L}=\underbrace{\|\mathbf{v}_{\theta}-(\mathbf{z}_{1}-\mathbf{z}_{0})\|^{2}}_{\text{transport fit(i)}}+\underbrace{\lambda\|\mathbf{v}_{\theta}-\mathbf{v}_{\text{RTE}}\|^{2}}_{\text{rte fit(ii) }}
\end{equation}
performs an \emph{L$^{2}$ projection}: minimising the second term encourages $\mathbf{v}_{\theta}$ to satisfy the diffusion--absorption balance locally, while the first term ensures the 
solution remains compatible with the clean target.
Consequently the ODE
\begin{equation}
\frac{d\mathbf{z}}{dt}=\mathbf{v}_{\theta}(\mathbf{z},t)
\end{equation}
no longer follows a bare straight line but a regularised trajectory whose tangent is the closest (in L$^{2}$) vector field that fulfils the RTE at every spatial index.
Mathematically, this embeds the continuous operator
\begin{equation}
(-D\nabla^{2}+\mu_{a}I)
\end{equation}
into the learning problem, yielding a physically regularised transport equation
\begin{equation}
\frac{d\mathbf{z}}{dt}=\Pi\!\left[(\mathbf{z}_{1}-\mathbf{z}_{0})\,\big|\,(-D\nabla^{2}+\mu_{a}I)\mathbf{z}_{t}\right],
\end{equation}
where $\Pi$ denotes the L$^{2}$ projection enforced by the combined loss. and $I$ is the identity operator acting point-wise on the latent feature map.
The final latent trajectory therefore satisfies both the data distribution and the latent-space analogue of RTE, producing dehazed outputs that are sharper and physically consistent while remaining end-to-end trainable.
%-------------------------------------------------------------------------

\begin{figure}[h!]
  \centering
  \includegraphics[width=1.0\linewidth]{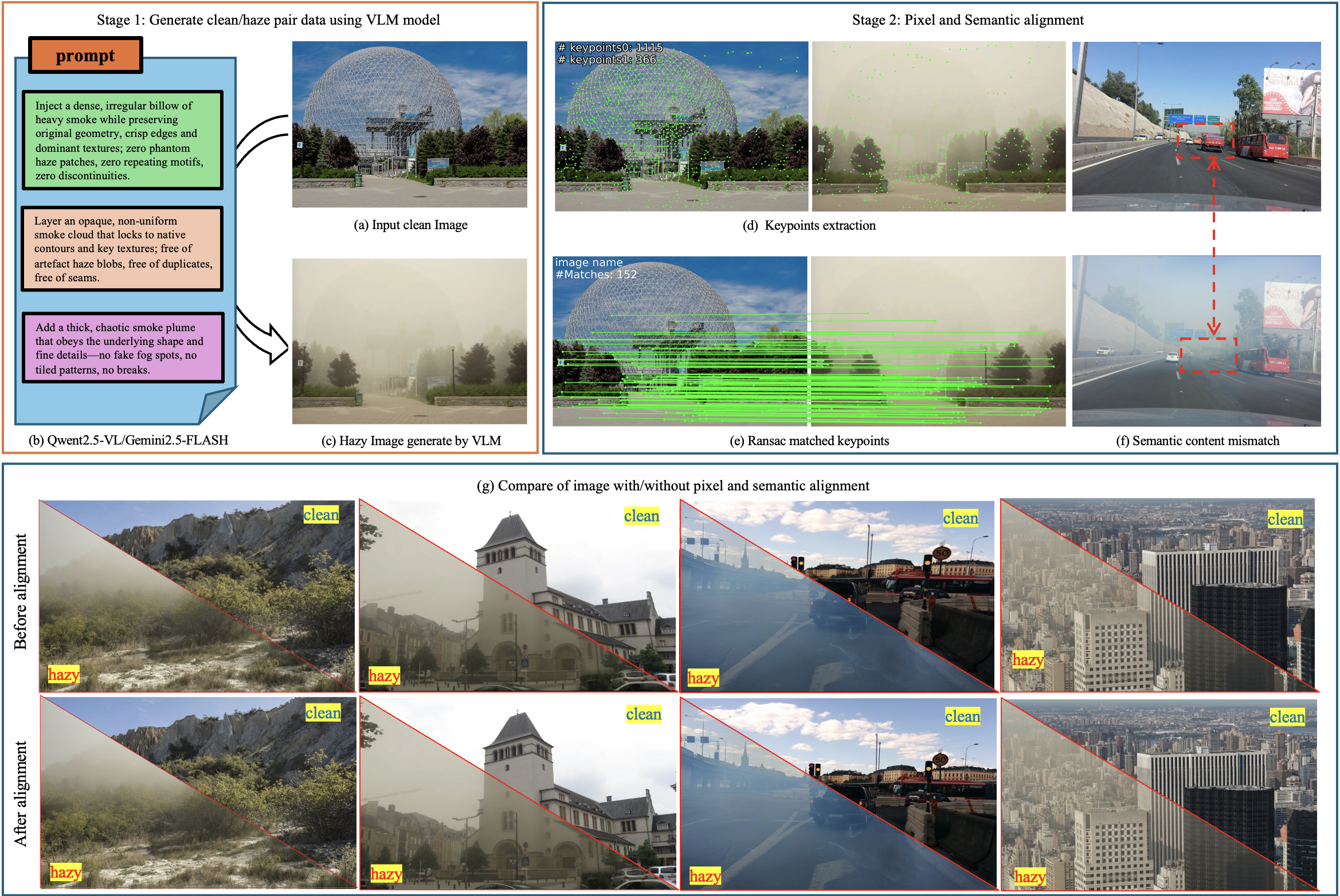}
  \caption{Overview of clean-to-haze image generation: Stage 1: conditioned on a text prompt, Qwen2-VL or Gemini-2-Flash edits the clean image to produce a realistic hazy image. Stage 2: dense keypoints are extracted from both images and used to estimate a homography that warps the hazy frame into pixel-wise alignment with the clean image, yielding the final training pair.}
  \label{fig:data}
\end{figure}

\subsection{Data Preparation}
Encouraged by recent advances in Vision-Language Models (VLMs), we therefore devise a scalable, scene-agnostic, hazy-image generation pipeline, illustrated in Figure~\ref{fig:data}.

\paragraph{\textbf{Stage 1: VLM-based hazy image generation.}}
Given a reference image, we first sample a stochastic haze prompt—two typical examples are:
\begin{itemize}
\item \textit{“Inject a dense, irregular billow of heavy smoke while preserving original geometry, crisp edges, and dominant textures; zero phantom haze patches, zero repeating motifs, zero discontinuities.”}
\item \textit{“Add a thick, chaotic smoke plume that obeys the underlying shape and fine details; no fake fog spots, no tiled patterns, no breaks.”}
\end{itemize}
The reference–prompt pair is then fed into Qwen2.5-VL~\cite{qwen2} or Gemini~\cite{gemini} to produce an initial hazy frame. Advanced VLMs implicitly capture spatial and geometric relationships, enabling the generated haze to exhibit density variations, depth-dependent attenuation, and realistic color shifts without requiring handcrafted depth or weather parameters.
Compared to ASM-based synthesis, VLM generation covers arbitrarily complex scenes and better reproduces non-stationary aerosol distributions—swirling smoke, localized fog pockets, etc. Because the process is prompt-driven, we can endlessly expand the training set by sampling new prompts, whereas ASM datasets remain constrained by curated depth maps and weather priors.
\paragraph{\textbf{Stage-2: Pixel alignment.}}
Although the hazy images produced by VLMs are visually realistic, their pixel coordinates are not accurately aligned with the corresponding clean inputs; most of them exhibit slight distortions that closely resemble those induced by changes in camera intrinsics.
To address this, we first extract dense keypoints on both clean and VLM-hazy frames and establish pixel-wise correspondences\cite{gim}.
These matches are then used to estimate a planar homography $\mathbf{H}\in\mathbb{R}^{3\times 3}$ with typically 800--1\,500 inliers.
Finally, the hazy frame is bilinear-warped onto the clean image via
\begin{equation}
I_{\text{aligned}}(\mathbf{x})=I_{\text{hazy}}\!\left(\mathbf{H}^{-1}\mathbf{x}\right),
\end{equation}
To suppress semantic artifacts (\eg, phantom objects or boundary distortions) in VLM outputs, we adopt a two-stage filtering strategy: automatic segmentation verification via SAM2\cite{ravi2024sam2} followed by manual human-in-the-loop inspection.

\paragraph{P-HAZE: A 50k Generate Dehazing Dataset.} 
Using this scalable pipeline, we produce P-HAZE, a large-scale dehazing dataset that comprises approximately 50\,000 rigorously aligned clear/haze pairs.
Reference images are gathered from Mapillary Vistas~\cite{mapillary}, Flickr2K~\cite{DIV2K+Flickr2K}, Unsplash2K\cite{Unsplash2K} and DIV2K~\cite{DIV2K+Flickr2K}, guaranteeing broad coverage of weather conditions, seasons, object categories, resolutions, and camera sensors; the resulting clear/hazy pairs are sufficiently diverse to support robust, real-world dehazer training.

\subsection{Training and Inference}
The training and inference procedures of RTE-FM-Dehazer are built on the flow matching framework to deliver an efficient, high-fidelity dehazing system.

%-------------------------------------------------------------------------
\begin{algorithm}[!htbp]
\caption{RTE-FM-Dehazer training step for one hazy--clean pair $(x_0,x_1)$.}
\label{alg:train}
\begin{algorithmic}[1]
\REQUIRE Hazy image $x_0$, clean image $x_1$, network parameters $\theta$, RTE-weight $\lambda$.
\ENSURE Updated parameters $\theta$.
\STATE Sample $t\sim\mathcal{U}(0,1)$
\STATE Interpolate state: $x_t \leftarrow (1-t)x_0 + t x_1$
\STATE Estimate coefficients: $(D,\mu_a)\leftarrow\phi_\theta(x_t)$
\STATE Compute Laplacian: $\nabla^2 x_t \leftarrow \text{LaplacianKernel}*x_t$
\STATE Build RTE velocity: $\hat{v}_{\text{RTE}}\leftarrow
       \text{RTE-velocity}(D,\mu_a,\nabla^2 x_t)$
\STATE Predict neural velocity: $v_\theta\leftarrow v_\theta(x_t,t)$
\STATE Compute losses:
\begin{ALC@g}
\STATE Flow matching loss: $\mathcal{L}_{\text{FM}}\leftarrow\|v_\theta-(x_1-x_0)\|_2^2$
\STATE RTE-consistency loss: $\mathcal{L}_{\text{RTE}}\leftarrow\|v_\theta-\hat{v}_{\text{RTE}}\|_F^2$
\STATE Total loss: $\mathcal{L}\leftarrow\mathcal{L}_{\text{FM}}+\lambda\mathcal{L}_{\text{RTE}}$
\end{ALC@g}
\STATE Back-propagate $\nabla_{\!\theta}\mathcal{L}$ and update $\theta$
\end{algorithmic}
\end{algorithm}

For every hazy–clear pair $(x_0,x_1)$ we sample $t\sim\mathcal U(0,1)$, build the linear interpolation $x_t=(1-t)x_0+t x_1$, and minimise the flow matching loss plus an RTE-consistency term (see Algorithm 1).  After training, the dehazing ODE $\frac{\mathrm d x}{\mathrm d t}=v_\theta(x,t)$ can be solved via $T$-step Euler integration:
\begin{equation}
x_1 \approx x_0 + \frac{1}{T}\sum_{i=0}^{T-1} v_\theta\!\left(x_{i/T}, \frac{i}{T}\right),
\end{equation}
which directly estimates the clean image from the hazy input using the vector field at t=0. While efficient, this approach may not fully capture the complexity of the image transformation.
To maximize visual quality, we recommend using more than 10 Euler steps. By subdividing the time interval [0, 1], the model more accurately simulates the gradual transition from hazy to clean images, producing higher-quality dehazed results, mitigating errors inherent to single-step approximations and better preserving the detail of the image and the color fidelity.

To reduce memory usage and accelerate training, we encode images into a compact latent space via the 8× autoencoder of Stable Diffusion v2.1 (SD2.1) \cite{SDXL}. This encoder downsamples images by a factor of 64 while preserving fine textures, enabling end-to-end latent training with 512×512 inputs on consumer-level GPUs.

\section{Experiments}

%-------------------------------------------------------------------------

\subsection{Experimental Settings}

We train solely on our synthetic P-HAZE dataset and evaluate generalization across five benchmarks that span distinct haze-generation mechanisms. I-HAZE\cite{I-HAZE} provides indoor fog-chamber images under uniform scattering. D-HAZE\cite{D-HAZE} contains suburban scenes filled with dense glycol fog, resulting in extremely low visibility. SMOKE\cite{SMOKE} comprises fire-smoke photographs produced with smoke generators. NH-HAZE\cite{NH-HAZE} supplies outdoor 4K sequences where depth-varying, non-uniform haze coexists with bright sky regions. RESIDE-6K\cite{reside} is a synthetic set with haze parameters derived from the accompanying depth maps.

We adopt PSNR (Peak Signal-to-Noise Ratio), SSIM (Structural Similarity Index)\cite{SSIM_PSNR}, and LPIPS (Learned Perceptual Image Patch Similarity)\cite{LPIPS} for evaluation. PSNR quantifies pixel-level fidelity, SSIM measures structural similarity (luminance, contrast, edges), and LPIPS evaluates deep perceptual distance; higher PSNR/SSIM and lower LPIPS indicate better quality.

\subsection{Implementation Details}
All training images are resized to $512\times512$.  
Latent codes are pre-computed to accelerate training.  
The model is trained for 200 epochs on 50\,000 synthetic hazy/clear pairs, consuming $\approx$\,20\,h on a single RTX~4090.  
Training adopts Adam with an initial learning rate of $1\mathrm{e}{-4}$, decayed by half every 50 epochs, and a batch size of 64.
We set the RTE weight to $\lambda=0.5$. Empirically, our method is insensitive to this choice, exhibiting stable performance across the range $[0.3, 0.7]$; we select $0.5$ as a balanced nominal value within this operational window.

\subsection{Network Architecture}
Our method builds upon a U-Net backbone adapted for latent-space RTE flow matching.
The framework comprises two main components:
\begin{enumerate}
    \item \textbf{Encoder--Decoder U-Net:}
    \begin{itemize}
        \item {Encoder:} four-level downsampling with ResNet-style blocks and spatial self-attention at the lowest resolution.
        \item {Decoder:} symmetric four-level upsampling with skip connections.
        \item {Bottleneck:} two ResBlocks with self-attention for global context.
    \end{itemize}
    \item \textbf{Latent Generator:} a pre-trained VAE encoder (SD~2.1) that maps hazy/clean images to a $64\times64\times4$ latent space, cutting computational cost while preserving perceptual quality.
\end{enumerate}

%---------------------main table----------------------------
\begin{table*}[b!]
\centering
\caption{Quantitative comparison on dehazing benchmarks. \textbf{Bold}: best; \underline{underlined}: second best. RTE-FM-Dehazer is trained solely on P-HAZE. We run the official code of every mentioned methods and evaluate with IQA-PyTorch~\cite{pyiqa}.}
\resizebox{\textwidth}{!}{
\begin{tabular}{|c|ccc|ccc|ccc|ccc|ccc|ccc|}
\toprule
\multirow{2}{*}{\large Methods} &
\multicolumn{3}{c|}{D-HAZE} &
\multicolumn{3}{c|}{I-HAZE} &
\multicolumn{3}{c|}{NH-HAZE} &
\multicolumn{3}{c|}{SMOKE} &
\multicolumn{3}{c|}{P-HAZE} &
\multicolumn{3}{c|}{RESIDE-6K} \\
\cmidrule{2-19}

& PSNR$\uparrow$ & SSIM$\uparrow$ & LPIPS$\downarrow$
& PSNR$\uparrow$ & SSIM$\uparrow$ & LPIPS$\downarrow$
& PSNR$\uparrow$ & SSIM$\uparrow$ & LPIPS$\downarrow$
& PSNR$\uparrow$ & SSIM$\uparrow$ & LPIPS$\downarrow$
& PSNR$\uparrow$ & SSIM$\uparrow$ & LPIPS$\downarrow$
& PSNR$\uparrow$ & SSIM$\uparrow$ & LPIPS$\downarrow$ \\
\midrule
DCP\cite{dark_channel} (CVPR2009)          & 11.38 & 0.438 & 0.704 & 11.56 & 0.542 & 0.355 & 12.04 & 0.515 & 0.505 & 11.75 & 0.488 & 0.405 & 16.87 & 0.762 & 0.303 & 17.89 & 0.867 & 0.097 \\
AODNet\cite{Aodnet} (ICCV2017)       &  8.17 & 0.316 & 0.873 &  9.85 & 0.525 & 0.487 &  9.31 & 0.328 & 0.700 & 10.26 & 0.394 & 0.570 &  9.43 & 0.470 & 0.505 &  9.35 & 0.599 & 0.324 \\
DehazeFormer\cite{dehazeformer} (TIP2023)  & 11.04 & 0.390 & 0.751 & 16.54 & 0.766 & 0.221 & 12.63 & 0.521 & 0.512 & 13.47 & 0.527 & 0.419 & 17.98 & 0.765 & 0.282 & \underline{31.42} & \textbf{0.981} & \underline{0.012} \\
RIDCP\cite{ridcp} (CVPR2023)        &  9.96 & \underline{0.474} & 0.618 & 16.71 & \textbf{0.817} & \underline{0.179} & 12.61 & \underline{0.631} & 0.417 & 13.23 & \underline{0.565} & 0.345 & 16.68 & 0.749 & 0.256 & 20.40 & 0.840 & 0.104 \\
C2PNet\cite{c2pnet} (CVPR2023)       & 11.22 & 0.388 & 0.761 & 16.51 & 0.745 & 0.247 & 12.59 & 0.519 & 0.516 & 13.72 & 0.532 & 0.420 & 16.84 & 0.735 & 0.293 & 24.90 & 0.937 & 0.034 \\
DEANet\cite{DEAnet} (TIP2024)        & 10.09 & 0.396 & 0.736 & \underline{16.96} & 0.766 & 0.220 & 12.38 & 0.535 & 0.503 & 13.68 & 0.540 & 0.406 & \underline{18.05} & \underline{0.786} & 0.270 & \textbf{31.78} & \underline{0.980} & \textbf{0.012} \\
IPC-Dehaze\cite{IPC-dehaze} (CVPR2025)   &  8.55 & 0.448 & 0.669 & 15.32 & 0.809 & 0.199 & 11.43 & \textbf{0.644} & 0.426 & 13.55 & \textbf{0.588} & \underline{0.336} & 16.02 & 0.771 & \underline{0.233} & 18.99 & 0.853 & 0.089 \\
LHTD\cite{LHTD} (CVPR2025)         & \textbf{12.84} & 0.465 & \underline{0.580} & 15.81 & 0.749 & 0.221 & \underline{15.60} & 0.543 & \underline{0.371} & \underline{15.67} & 0.406 & 0.355 & 13.98 & 0.614 & 0.288 & 17.19 & 0.711 & 0.198 \\
\midrule
RTE-FM-Dehazer (Ours)    & \underline{12.56} & \textbf{0.480} & \textbf{0.528} & \textbf{18.81} & \underline{0.812} & \textbf{0.145} & \textbf{19.53} & 0.600 & \textbf{0.233} & \textbf{18.67} & 0.554 & \textbf{0.222} & \textbf{20.29} & \textbf{0.792} & \textbf{0.159} & 21.77 & 0.858 & 0.068 \\
\bottomrule
\end{tabular}}

\label{tab:dehazing}
\end{table*}

\subsection{Comparisons with State-of-the-art Methods}

\begin{figure*}[h!]
  \centering
  \includegraphics[width=1.0\linewidth]{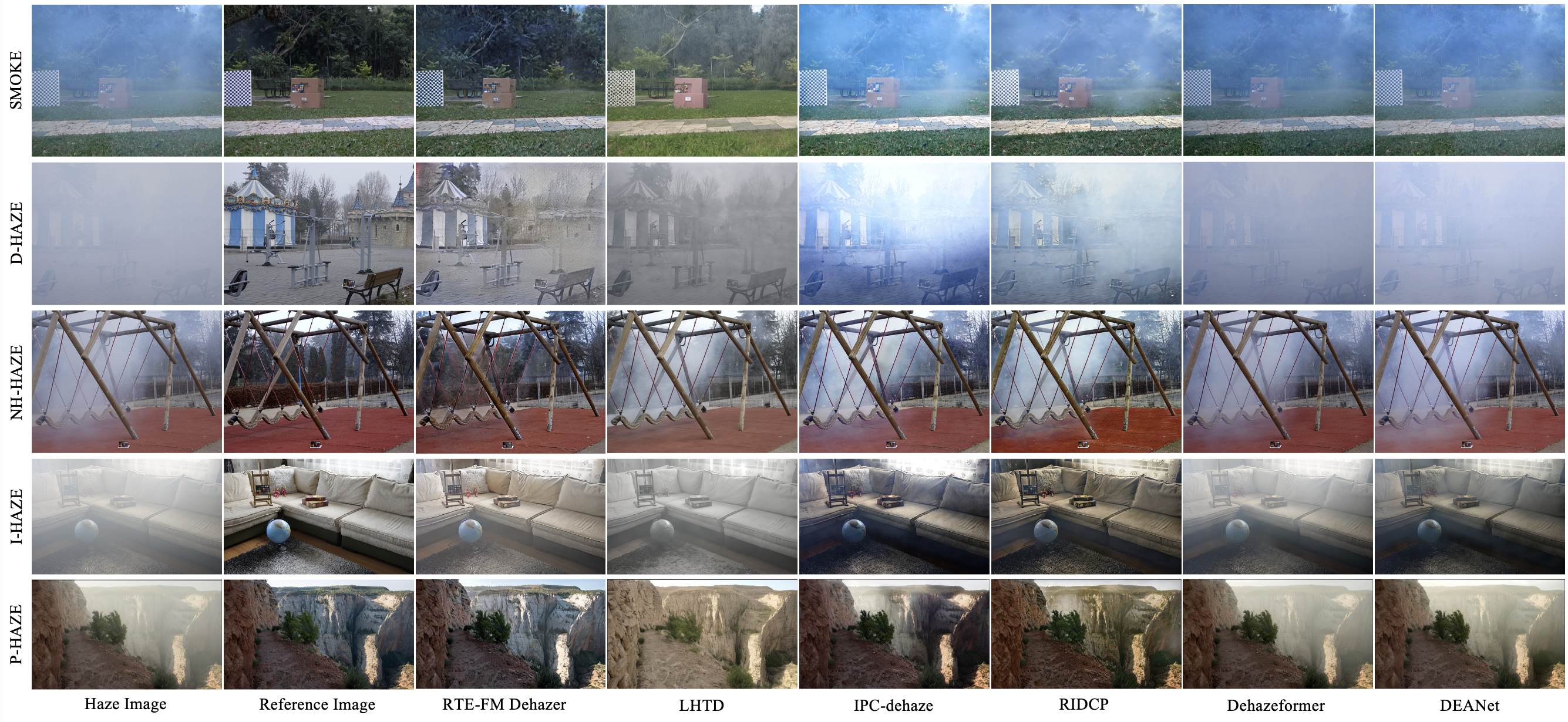}
  \caption{Visual comparison on challenging hazy benchmarks. Zoom in for best detail.}
  \label{fig:showresult}
\end{figure*}

\paragraph{Quantitative Evaluation.}  
We conduct quantitative comparisons on six public dehazing benchmarks that jointly cover synthetic, real indoor, real outdoor, and smoke-laden hazy scenes.
As reported in Table~\ref{tab:dehazing}, RTE-FM-Dehazer achieves the highest PSNR on four datasets and the lowest LPIPS on five out of six, demonstrating superior pixel fidelity and perceptual quality.
In detail, on the challenging NH-HAZE dataset, we outperform the strongest competitor LHTD by +3.9 dB in PSNR and reduce LPIPS by 37\%; on the SMOKE dataset, the gain remains +3.0 dB in PSNR and --37\% in LPIPS.
Although DehazeFormer and DEANet achieve higher SSIM on the synthetic RESIDE-6K, their PSNR drops by more than 9 dB on the four real-captured benchmarks, confirming the stronger generalisation of RTE-FM-Dehazer to real-world haze.

\begin{figure}[h!]
  \centering
  \includegraphics[width=1.0\linewidth]{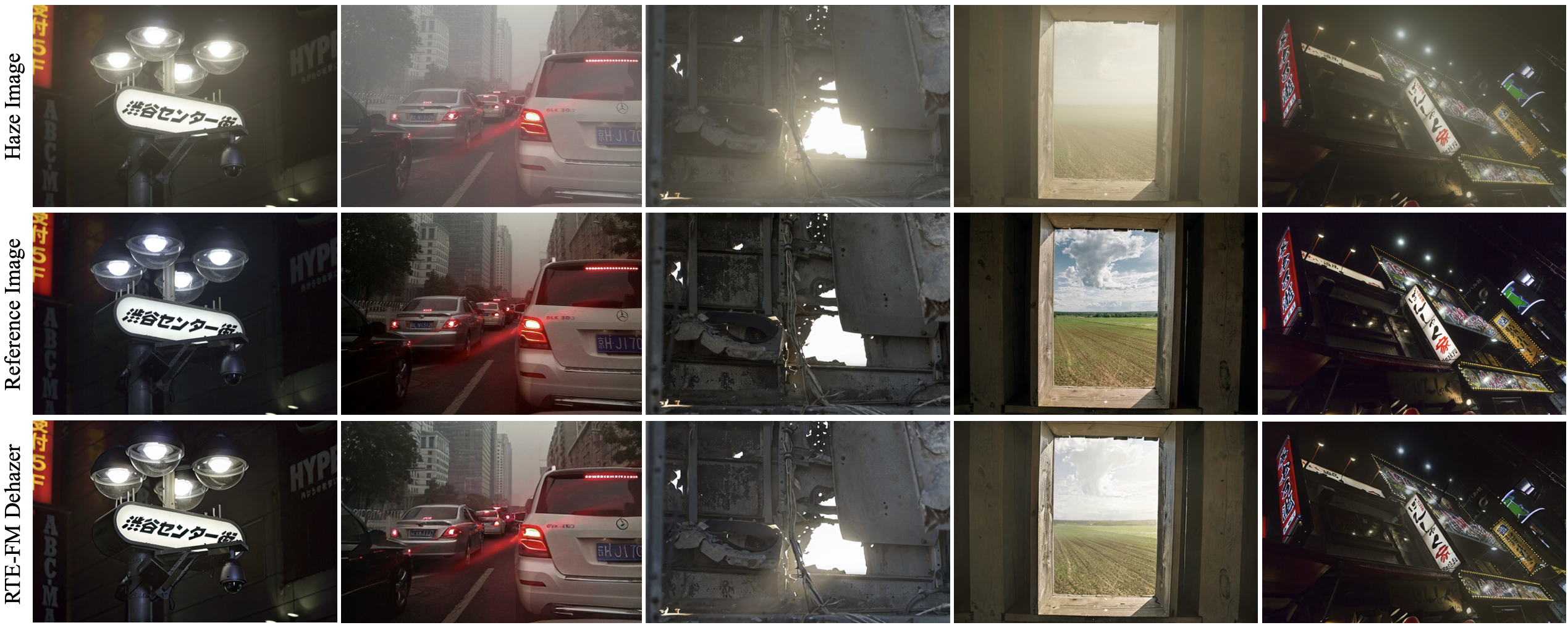}
  \caption{Visual dehazing results under direct, scattered, and reflected lighting. RTE-FM-Dehazer effectively removes haze while preserving diverse lighting structures.}
  \label{fig:light transport}
\end{figure}

\begin{figure}[!h]
  \centering
  \includegraphics[width=\linewidth]{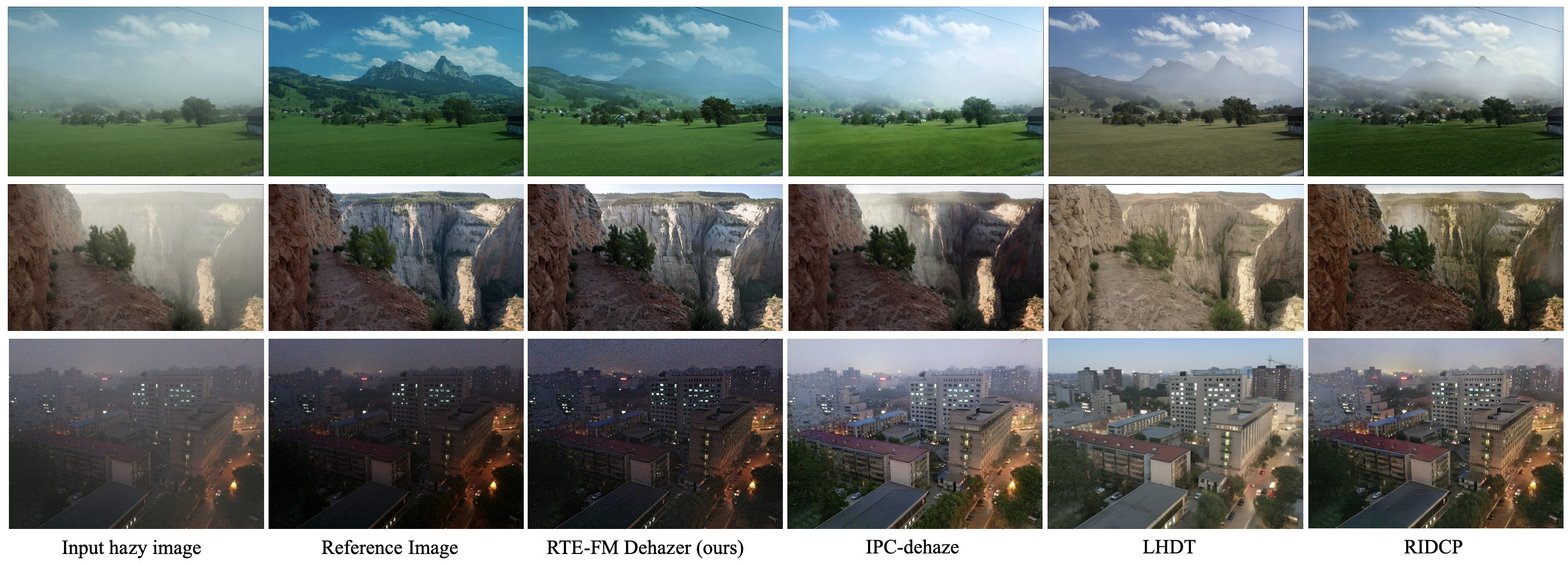}
  \caption{Compared with recent SOTA generative-prior methods, RTE-FM-Dehazer provides color restoration that matches the ground-truth colors more closely.}
  \label{fig:color shift}
\end{figure}

\paragraph{Qualitative Evaluation.}  
Visual comparisons on real hazy images are shown in Figure~\ref{fig:showresult}. CNN-based and transformer-based methods (DehazeFormer, DEA-Net) leave residual haze, especially in dense regions. VQGAN priors (RIDCP, IPC-Dehaze) tend to over-expose and introduce color casts. GAN-based LHTD produces competitive dehazing but suffers from loss of detail and patch-like artifacts. In contrast, RTE-FM-Dehazer removes haze of varying density while preserving texture and color. Even under extreme haze it recovers images that are noise-suppressed and artifact-reduced, and when dust or foreground objects overlap the haze, it correctly removes turbidity without retaining floating particles. Figure~\ref{fig:light transport} demonstrates that the original light paths and global color distribution are preserved, guaranteeing chromatic consistency across varying lighting conditions. Figure~\ref{fig:color shift} shows that our method corrects color shifts caused by low-light or particle pollution, producing visually faithful color.  These qualitative results corroborate the quantitative rankings and verify that RTE-FM-Dehazer consistently generates high-quality, color-consistent, natural haze-free images in diverse real-world scenarios.

\begin{table}[b!]
\centering
\caption{Ablation study on regularizer components and training data. RTE-FM outperforms decoupled regularizer baselines.}
\label{tab:ablation}
\resizebox{\textwidth}{!}{%
\begin{tabular}{|c|ccc|ccc|ccc|ccc|ccc|ccc|}
\toprule
\multirow{2}{*}{Methods} &
\multicolumn{3}{c}{P-HAZE} &
\multicolumn{3}{c}{I-HAZE} &
\multicolumn{3}{c}{NH-HAZE} &
\multicolumn{3}{c}{D-HAZE} &
\multicolumn{3}{c}{SMOKE} &
\multicolumn{3}{c}{RESIDE-6K} \\
\cmidrule(lr){2-4} \cmidrule(lr){5-7} \cmidrule(lr){8-10} \cmidrule(lr){11-13} \cmidrule(lr){14-16} \cmidrule(lr){17-19}
 & PSNR$\uparrow$ & SSIM$\uparrow$ & LPIPS$\downarrow$ & PSNR$\uparrow$ & SSIM$\uparrow$ & LPIPS$\downarrow$ & PSNR$\uparrow$ & SSIM$\uparrow$ & LPIPS$\downarrow$ & PSNR$\uparrow$ & SSIM$\uparrow$ & LPIPS$\downarrow$ & PSNR$\uparrow$ & SSIM$\uparrow$ & LPIPS$\downarrow$ & PSNR$\uparrow$ & SSIM$\uparrow$ & LPIPS$\downarrow$ \\
\midrule

Absorb-only FM & 18.11 & 0.636 & 0.378 & 11.27 & 0.498 & 0.311 & 12.77 & 0.459 & 0.441 & 9.29 & 0.330 & 0.571 & 11.88 & 0.400 & 0.399 & 16.77 & 0.793 & 0.165 \\
Diffusion-only FM & 18.91 & 0.552 & 0.356 & 11.19 & 0.577 & 0.398 & 13.85 & 0.505 & 0.480 & 9.11 & 0.324 & 0.601 & 12.21 & 0.421 & 0.411 & 12.03 & 0.693 & 0.201 \\
DCP-regularizer-FM & 17.22 & 0.621 & 0.344 & 12.21 & 0.455 & 0.444 & 12.22 & 0.523 & 0.547 & 9.99 & 0.369 & 0.621 & 11.19 & 0.398 & 0.412 & 11.79 & 0.777 & 0.189 \\
Rectified FM & 19.21 & 0.768 & 0.254 & 17.61 & 0.762 & 0.193 & 12.99 & 0.494 & 0.491 & 11.07 & 0.426 & 0.681 & 13.40 & 0.463 & 0.372 & 16.59 & 0.746 & 0.208 \\
Vanilla FM & 20.12 & 0.785  & 0.159 & 18.27 & 0.805 & 0.162 & 13.78 & 0.507  & 0.406  & 11.43  & 0.453  & 0.614  & 13.77  & 0.488  & 0.351  & 17.51 & 0.783 & 0.112 \\

\midrule
RTE-FM(Ours) & 20.29 & 0.792 & 0.159 & 18.81 & 0.812 & 0.145 & 19.53 & 0.600 & 0.233 & 12.56 & 0.480 & 0.528 & 18.67 & 0.554 & 0.222 & 21.77 & 0.858 & 0.068 \\

\bottomrule
\end{tabular}}
\end{table}

\subsection{Ablation on Regularizer Components}
\begin{figure}[b!]
  \centering
  \includegraphics[width=\linewidth]{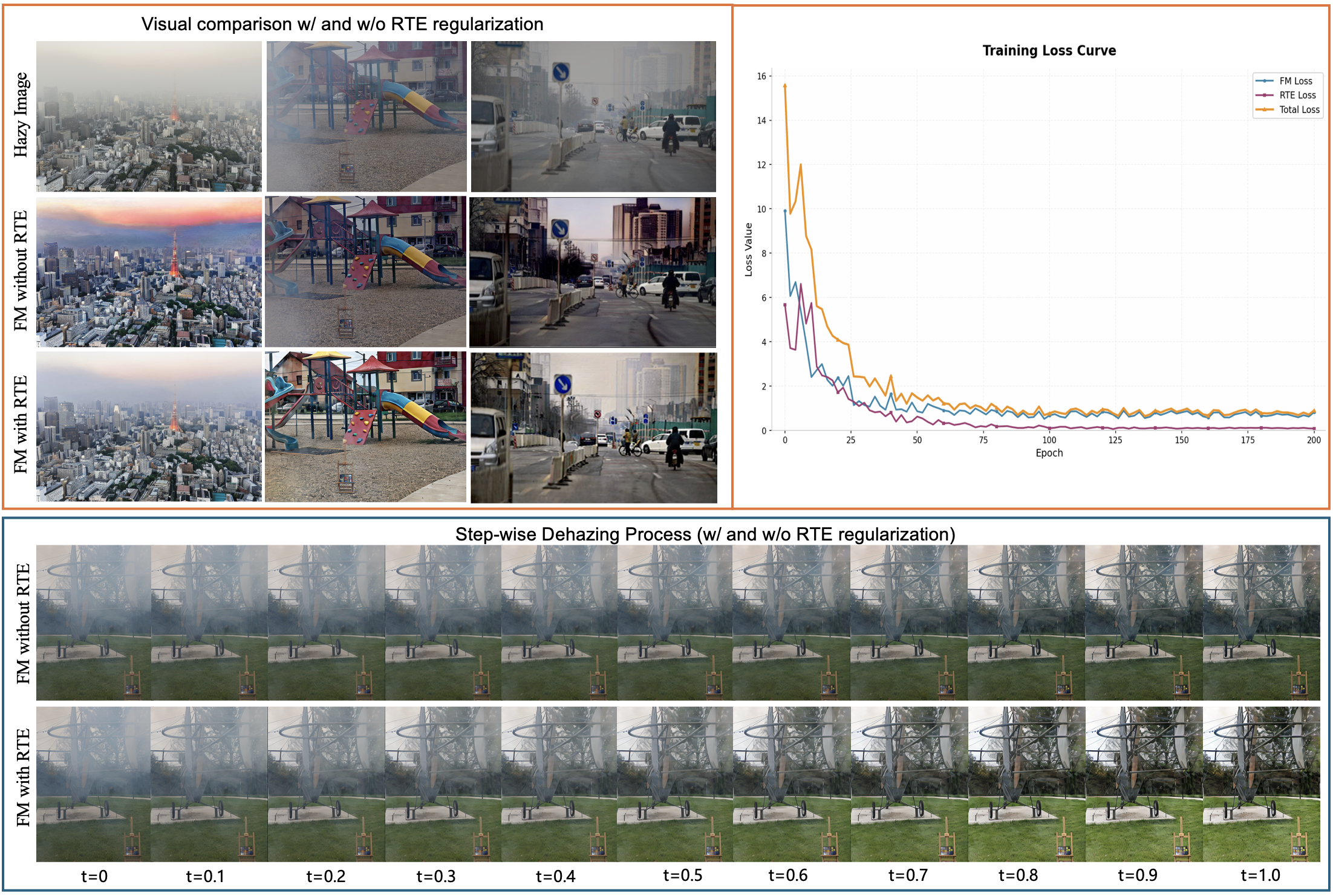}
  \caption{Ablation study results comparing the performance of the proposed FM model with different regularization.}
  \label{fig:ablation_rte}
\end{figure}

To isolate the necessity of RTE regularization, we train four model variants on P-HAZE: Vanilla FM without RTE regularization ($\lambda=0$), Absorb-only FM using only the absorption term $-\mu_a\mathbf{z}$, Diffusion-only FM using only the Laplacian term $D\nabla^2\mathbf{z}$, and DCP-regularized FM replacing RTE with dark-channel prior loss that explicitly enforces the loss gradient to align with dark channel minimization. Rectified Flow\cite{rectifiedflow} learns geometrically straight-line paths in a data-driven manner, while RTE-FM  constrains these trajectories with RTE to ensure optical consistency. Table~\ref{tab:ablation} reveals that decoupled regularization catastrophically fails on non-uniform haze due to incomplete atmospheric modeling. Absorb-only suffers from severe performance degradation and blotchy artifacts due to the lack of spatial smoothing, while Diffusion-only leaves significant residual haze and fails to recover clear structures without attenuation guidance.Notably, DCP-regularized FM underperforms even Vanilla FM, indicating that heuristic hard constraints fundamentally misalign with flow matching's transport cost optimization. In contrast, RTE-FM's integrated absorption-diffusion formulation provides a compatible differentiable constraint that preserves generative flexibility while enforcing radiative fidelity. This superior inductive bias enables robust cross-domain generalization (NH-HAZE, SMOKE) and prevents overfitting to synthetic patterns. Training curves (Fig.~\ref{fig:ablation_rte}) confirm RTE loss dominates early training, stabilizing optimization against local optima and ensuring monotonic radiance recovery, unlike Vanilla FM's erratic early-timestep artifacts.

\section{Limitation and Conclusion}

RTE-FM-Dehazer combines flow matching's macroscopic transport with the microscopic Radiative Transfer Equation: the learned velocity field is projected onto RTE-derived energy flux, guiding a single, deterministic trajectory from hazy to clean latent space without stochastic sampling or post-processing. Extensive benchmarks demonstrate effective removal of non-uniform haze, smoke, and color casts while preserving fine textures. 

Moving forward, we will pursue methods to address remaining constraints. Operating directly on pixel space will eliminate the information loss inherent in VAE latent encoding, enabling fine-grained detail preservation and improved robustness in ultra-dense haze scenarios where current latent approximations produce artifacts. Furthermore, we will investigate efficient numerical solvers, progressive sampling strategies, and model distillation techniques to ensure computational feasibility at full resolution without sacrificing the physical consistency of the dehazing process.

%\section*{Acknowledgements}
%Please insert your acknowledgments here.

\bibliographystyle{splncs04}
\bibliography{main}

@String(BMVC  = {Brit. Mach. Vis. Conf.})

@String(CVPRW = {IEEE Conf. Comput. Vis. Pattern Recog. Worksh.})

@String(AAAI  = {AAAI})

@String(ICIP  = {IEEE Int. Conf. Image Process.})

@String(TOG   = {ACM Trans. Graph.})

@String(BMVC  =	{BMVC})

@String(CVPRW = {CVPRW})

@String(ICIP  = {ICIP})

@String(TOG   = {ACM TOG})

@article{ASM,
  title={Optics of the atmosphere: scattering by molecules and particles},
  author={McCartney, Earl J},
  journal={New york},
  year={1976}
}

@inbook{RTE,
  author    = {Zhao, C.},
  title     = {Numerical Methods for Solving the Radiative Transfer Equation},
  booktitle = {Thermal Radiation: From Macro to Nano},
  pages     = {54--79},
  year      = {2024},
  publisher = {Cambridge University Press}
}

@article{dark_channel,
  title={Single image haze removal using dark channel prior},
  author={He, Kaiming and Sun, Jian and Tang, Xiaoou},
  journal={IEEE transactions on pattern analysis and machine intelligence},
  volume={33},
  number={12},
  pages={2341--2353},
  year={2010},
  publisher={IEEE}
}

@article{color_lines,
  title={Dehazing using color-lines},
  author={Fattal, Raanan},
  journal={ACM transactions on graphics (TOG)},
  volume={34},
  number={1},
  pages={1--14},
  year={2014},
  publisher={ACM New York, NY, USA}
}

@inproceedings{color_attenuation,
  title={Single image dehazing using color attenuation prior.},
  author={Zhu, Qingsong and Mai, Jiaming and Shao, Ling},
  booktitle={BMVC},
  volume={4},
  pages={1674--1682},
  year={2014}
}

@article{haze_line,
  author={Berman, Dana and Treibitz, Tali and Avidan, Shai},
  journal={IEEE Transactions on Pattern Analysis and Machine Intelligence}, 
  title={Single Image Dehazing Using Haze-Lines}, 
  year={2020},
  volume={42},
  number={3},
  pages={720-734}
}

@article{Dehazenet,
  title={Dehazenet: An end-to-end system for single image haze removal},
  author={Cai, Bolun and Xu, Xiangmin and Jia, Kui and Qing, Chunmei and Tao, Dacheng},
  journal={IEEE transactions on image processing},
  volume={25},
  number={11},
  pages={5187--5198},
  year={2016},
  publisher={IEEE}
}

@inproceedings{densely_dehaze,
  title={Densely connected pyramid dehazing network},
  author={Zhang, He and Patel, Vishal M},
  booktitle={Proceedings of the IEEE conference on computer vision and pattern recognition},
  pages={3194--3203},
  year={2018}
}

@inproceedings{Aodnet,
  title={Aod-net: All-in-one dehazing network},
  author={Li, Boyi and Peng, Xiulian and Wang, Zhangyang and Xu, Jizheng and Feng, Dan},
  booktitle={Proceedings of the IEEE international conference on computer vision},
  pages={4770--4778},
  year={2017}
}

@inproceedings{FFAnet,
  title={FFA-Net: Feature fusion attention network for single image dehazing},
  author={Qin, Xu and Wang, Zhilin and Bai, Yuanchao and Xie, Xiaodong and Jia, Huizhu},
  booktitle={Proceedings of the AAAI conference on artificial intelligence},
  volume={34},
  number={07},
  pages={11908--11915},
  year={2020}
}

@article{DEAnet,
  title={DEA-Net: Single image dehazing based on detail-enhanced convolution and content-guided attention},
  author={Chen, Zixuan and He, Zewei and Lu, Zhe-Ming},
  journal={IEEE transactions on image processing},
  volume={33},
  pages={1002--1015},
  year={2024},
  publisher={IEEE}
}

@inproceedings{dehazeXL,
  title={Tokenize image patches: Global context fusion for effective haze removal in large images},
  author={Chen, Jiuchen and Yan, Xinyu and Xu, Qizhi and Li, Kaiqi},
  booktitle={Proceedings of the Computer Vision and Pattern Recognition Conference},
  pages={2258--2268},
  year={2025}
}

@inproceedings{dehaze3D,
  title={Image dehazing transformer with transmission-aware 3d position embedding},
  author={Guo, Chun-Le and Yan, Qixin and Anwar, Saeed and Cong, Runmin and Ren, Wenqi and Li, Chongyi},
  booktitle={Proceedings of the IEEE/CVF conference on computer vision and pattern recognition},
  pages={5812--5820},
  year={2022}
}

@inproceedings{domain_dehazing,
  title={Domain adaptation for image dehazing},
  author={Shao, Yuanjie and Li, Lerenhan and Ren, Wenqi and Gao, Changxin and Sang, Nong},
  booktitle={Proceedings of the IEEE/CVF conference on computer vision and pattern recognition},
  pages={2808--2817},
  year={2020}
}

@article{unsupervised_dark_channel,
  title={Unsupervised single image dehazing using dark channel prior loss},
  author={Golts, Alona and Freedman, Daniel and Elad, Michael},
  journal={IEEE transactions on Image Processing},
  volume={29},
  pages={2692--2701},
  year={2019},
  publisher={IEEE}
}

@article{ucl-dehaze,
  title={UCL-dehaze: Toward real-world image dehazing via unsupervised contrastive learning},
  author={Wang, Yongzhen and Yan, Xuefeng and Wang, Fu Lee and Xie, Haoran and Yang, Wenhan and Zhang, Xiao-Ping and Qin, Jing and Wei, Mingqiang},
  journal={IEEE Transactions on Image Processing},
  volume={33},
  pages={1361--1374},
  year={2024},
  publisher={IEEE}
}

@article{refinednet,
  title={RefineDNet: A weakly supervised refinement framework for single image dehazing},
  author={Zhao, Shiyu and Zhang, Lin and Shen, Ying and Zhou, Yicong},
  journal={IEEE Transactions on Image Processing},
  volume={30},
  pages={3391--3404},
  year={2021},
  publisher={IEEE}
}

@inproceedings{condition_dehazing,
  title={Single image dehazing via conditional generative adversarial network},
  author={Li, Runde and Pan, Jinshan and Li, Zechao and Tang, Jinhui},
  booktitle={Proceedings of the IEEE conference on computer vision and pattern recognition},
  pages={8202--8211},
  year={2018}
}

@inproceedings{ridcp,
  title={Ridcp: Revitalizing real image dehazing via high-quality codebook priors},
  author={Wu, Rui-Qi and Duan, Zheng-Peng and Guo, Chun-Le and Chai, Zhi and Li, Chongyi},
  booktitle={Proceedings of the IEEE/CVF conference on computer vision and pattern recognition},
  pages={22282--22291},
  year={2023}
}

@inproceedings{IPC-dehaze,
  title={Iterative Predictor-Critic Code Decoding for Real-World Image Dehazing},
  author={Fu, Jiayi and Liu, Siyu and Liu, Zikun and Guo, Chun-Le and Park, Hyunhee and Wu, Ruiqi and Wang, Guoqing and Li, Chongyi},
  booktitle={Proceedings of the Computer Vision and Pattern Recognition Conference},
  pages={12700--12709},
  year={2025}
}

@inproceedings{dehazediff,
  title={DehazeDiff: When conditional guidance meets diffusion models for image dehazing},
  author={Cheng, Longyu and Ba, Xujin and Qu, Yanyun},
  booktitle={2024 IEEE International Symposium on Circuits and Systems (ISCAS)},
  pages={1--5},
  year={2024},
  organization={IEEE}
}

@inproceedings{LHTD,
  title={Learning hazing to dehazing: Towards realistic haze generation for real-world image dehazing},
  author={Wang, Ruiyi and Zheng, Yushuo and Zhang, Zicheng and Li, Chunyi and Liu, Shuaicheng and Zhai, Guangtao and Liu, Xiaohong},
  booktitle={Proceedings of the Computer Vision and Pattern Recognition Conference},
  pages={23091--23100},
  year={2025}
}

@inproceedings{c2pnet,
  title={Curricular Contrastive Regularization for Physics-aware Single Image Dehazing},
  author={Zheng, Yu and Zhan, Jiahui and He, Shengfeng and Dong, Junyu and Du, Yong},
  booktitle={IEEE/CVF Conference on Computer Vision and Pattern Recognition},
  year={2023}
}

@article{dehazeformer,
  title={Vision Transformers for Single Image Dehazing},
  author={Song, Yuda and He, Zhuqing and Qian, Hui and Du, Xin},
  journal={IEEE Transactions on Image Processing},
  year={2023},
  volume={32},
  pages={1927-1941}
}

@article{VQGAN,
  title={Vector-quantized image modeling with improved vqgan},
  author={Yu, Jiahui and Li, Xin and Koh, Jing Yu and Zhang, Han and Pang, Ruoming and Qin, James and Ku, Alexander and Xu, Yuanzhong and Baldridge, Jason and Wu, Yonghui},
  journal={arXiv preprint arXiv:2110.04627},
  year={2021}
}

@article{VQVAE,
  title={Generating diverse high-fidelity images with vq-vae-2},
  author={Razavi, Ali and Van den Oord, Aaron and Vinyals, Oriol},
  journal={Advances in neural information processing systems},
  volume={32},
  year={2019}
}

@article{flow_matching,
  title={Flow matching for generative modeling},
  author={Lipman, Yaron and Chen, Ricky TQ and Ben-Hamu, Heli and Nickel, Maximilian and Le, Matt},
  journal={arXiv preprint arXiv:2210.02747},
  year={2022}
}

@article{qwen2,
  title={Qwen2. 5-vl technical report},
  author={Bai, Shuai and Chen, Keqin and Liu, Xuejing and Wang, Jialin and Ge, Wenbin and Song, Sibo and Dang, Kai and Wang, Peng and Wang, Shijie and Tang, Jun and others},
  journal={arXiv preprint arXiv:2502.13923},
  year={2025}
}

@article{gemini,
  title={Gemini: a family of highly capable multimodal models},
  author={Team, Gemini and Anil, Rohan and Borgeaud, Sebastian and Alayrac, Jean-Baptiste and Yu, Jiahui and Soricut, Radu and Schalkwyk, Johan and Dai, Andrew M and Hauth, Anja and Millican, Katie and others},
  journal={arXiv preprint arXiv:2312.11805},
  year={2023}
}

@article{diffusion,
  title={Diffusion models in vision: A survey},
  author={Croitoru, Florinel-Alin and Hondru, Vlad and Ionescu, Radu Tudor and Shah, Mubarak},
  journal={IEEE transactions on pattern analysis and machine intelligence},
  volume={45},
  number={9},
  pages={10850--10869},
  year={2023},
  publisher={Ieee}
}

@article{SDXL,
  title={Sdxl: Improving latent diffusion models for high-resolution image synthesis},
  author={Podell, Dustin and English, Zion and Lacey, Kyle and Blattmann, Andreas and Dockhorn, Tim and M{\"u}ller, Jonas and Penna, Joe and Rombach, Robin},
  journal={arXiv preprint arXiv:2307.01952},
  year={2023}
}

@article{ravi2024sam2,
  title={SAM 2: Segment Anything in Images and Videos},
  author={Ravi, Nikhila and Gabeur, Valentin and Hu, Yuan-Ting and Hu, Ronghang and Ryali, Chaitanya and Ma, Tengyu and Khedr, Haitham and R{\"a}dle, Roman and Rolland, Chloe and Gustafson, Laura and Mintun, Eric and Pan, Junting and Alwala, Kalyan Vasudev and Carion, Nicolas and Wu, Chao-Yuan and Girshick, Ross and Doll{\'a}r, Piotr and Feichtenhofer, Christoph},
  journal={arXiv preprint arXiv:2408.00714},
  year={2024}
}

@article{rectifiedflow,
  title={Flow straight and fast: Learning to generate and transfer data with rectified flow},
  author={Liu, Xingchao and Gong, Chengyue and Liu, Qiang},
  journal={arXiv preprint arXiv:2209.03003},
  year={2022}
}

@article{reside,
  title={Benchmarking single-image dehazing and beyond},
  author={Li, Boyi and Ren, Wenqi and Fu, Dengpan and Tao, Dacheng and Feng, Dan and Zeng, Wenjun and Wang, Zhangyang},
  journal={IEEE transactions on image processing},
  volume={28},
  number={1},
  pages={492--505},
  year={2018},
  publisher={IEEE}
}

@inproceedings{haze4k,
  title={Pixel transformer for synthetic-to-real single image dehazing},
  author={Zhang, Yuting and Wang, Fan and Yin, Dong},
  booktitle={2023 8th International Conference on Communication, Image and Signal Processing (CCISP)},
  pages={250--254},
  year={2023},
  organization={IEEE}
}

@inproceedings{NH-haze,
  title={NH-HAZE: An image dehazing benchmark with non-homogeneous hazy and haze-free images},
  author={Ancuti, Codruta O and Ancuti, Cosmin and Timofte, Radu},
  booktitle={Proceedings of the IEEE/CVF conference on computer vision and pattern recognition workshops},
  pages={444--445},
  year={2020}
}

@inproceedings{D-haze,
  title={D-hazy: A dataset to evaluate quantitatively dehazing algorithms},
  author={Ancuti, Cosmin and Ancuti, Codruta O and De Vleeschouwer, Christophe},
  booktitle={2016 IEEE international conference on image processing (ICIP)},
  pages={2226--2230},
  year={2016},
  organization={IEEE}
}

@inproceedings{I-haze,
  title={I-HAZE: A dehazing benchmark with real hazy and haze-free indoor images},
  author={Ancuti, Cosmin and Ancuti, Codruta O and Timofte, Radu and De Vleeschouwer, Christophe},
  booktitle={International conference on advanced concepts for intelligent vision systems},
  pages={620--631},
  year={2018},
  organization={Springer}
}

@inproceedings{SMOKE,
  title={Structure representation network and uncertainty feedback learning for dense non-uniform fog removal},
  author={Jin, Yeying and Yan, Wending and Yang, Wenhan and Tan, Robby T},
  booktitle={Asian Conference on Computer Vision},
  pages={155--172},
  year={2022},
  organization={Springer}
}

@inproceedings{SSIM_PSNR,
  title={Image quality metrics: PSNR vs. SSIM},
  author={Hore, Alain and Ziou, Djemel},
  booktitle={2010 20th international conference on pattern recognition},
  pages={2366--2369},
  year={2010},
  organization={IEEE}
}

@inproceedings{LPIPS,
  title={The unreasonable effectiveness of deep features as a perceptual metric},
  author={Zhang, Richard and Isola, Phillip and Efros, Alexei A and Shechtman, Eli and Wang, Oliver},
  booktitle={Proceedings of the IEEE conference on computer vision and pattern recognition},
  pages={586--595},
  year={2018}
}

@misc{pyiqa,
  title={{IQA-PyTorch}: PyTorch Toolbox for Image Quality Assessment},
  author={Chaofeng Chen and Jiadi Mo},
  year={2022}
}

@article{gim,
  title={Gim: Learning generalizable image matcher from internet videos},
  author={Shen, Xuelun and Cai, Zhipeng and Yin, Wei and M{\"u}ller, Matthias and Li, Zijun and Wang, Kaixuan and Chen, Xiaozhi and Wang, Cheng},
  journal={arXiv preprint arXiv:2402.11095},
  year={2024}
}

@inproceedings{DIV2K+Flickr2K,
  title={Ntire 2017 challenge on single image super-resolution: Dataset and study},
  author={Agustsson, Eirikur and Timofte, Radu},
  booktitle={Proceedings of the IEEE conference on computer vision and pattern recognition workshops},
  pages={126--135},
  year={2017}
}

@article{Unsplash2K,
  title={Noise Conditional Flow Model for Learning the Super-Resolution Space},
  author={Younggeun Kim and Donghee Son},
  journal={2021 IEEE/CVF Conference on Computer Vision and Pattern Recognition Workshops (CVPRW)},
  year={2021},
  pages={424-432}
}

@inproceedings{mapillary,
  title={The mapillary vistas dataset for semantic understanding of street scenes},
  author={Neuhold, Gerhard and Ollmann, Tobias and Rota Bulo, Samuel and Kontschieder, Peter},
  booktitle={Proceedings of the IEEE international conference on computer vision},
  pages={4990--4999},
  year={2017}
}
\end{document}